\PassOptionsToPackage{numbers,sort&compress}{natbib}
\documentclass{article}

\usepackage[final]{mlwg_2019}

\usepackage[utf8]{inputenc} 
\usepackage[T1]{fontenc}    
\usepackage{hyperref}       
\usepackage{url}            
\usepackage{booktabs}       
\usepackage{amsfonts}       
\usepackage{nicefrac}       
\usepackage{microtype}      

\usepackage{graphicx}
\usepackage{subcaption}
\usepackage{caption}
\usepackage{amsmath}
\usepackage{amssymb}
\DeclareMathOperator*{\argmax}{argmax}

\title{Neural Network Memorization Dissection}

\author{
Jindong Gu \\
The University of Munich\\
Siemens AG, Corporate Technology\\
\texttt{jindong.gu@siemens.com} \\
\And
Volker Tresp \\
The University of Munich\\
Siemens AG, Corporate Technology\\
\texttt{volker.tresp@siemens.com} \\
}

\begin{document}

\maketitle
\vspace{-0.5cm}

\begin{abstract}
Deep neural networks (DNNs) can easily fit a random labeling of the training data with zero training error. What is the difference between DNNs trained with random labels and the ones trained with true labels? Our paper answers this question with two contributions. First, we study the memorization properties of DNNs. Our empirical experiments shed light on how DNNs prioritize the learning of simple input patterns. In the second part, we propose to measure the similarity between what different DNNs have learned and memorized. With the proposed approach, we analyze and compare DNNs trained on data with true labels and random labels. The analysis shows that DNNs have \textit{One way to Learn} and \textit{N ways to Memorize}. We also use gradient information to gain an understanding of the analysis results.
\end{abstract}

\section{Introduction}
\vspace{-0.1cm}
From a traditional perspective of machine learning, a model with sufficient capacity can overfit the training data, e.g., by memorizing each training sample. The overfitted model will yield poor performance on unseen test data \cite{GoodfellowDL}. However, over-parametrized deep neural networks (DNNs) with high expressiveness often show excellent generalization ability, even without explicit regularization techniques. \cite{Zhang2016UnderstandingDL} showed that the generalization of DNNs cannot be explained by traditional approaches, such as \textit{Rademacher complexity} \cite{Bartlett2001RademacherAG}, \textit{VC-dimension} \cite{Vapnik1998}, and \textit{Uniform stability} \cite{bousquet2002stability}).

\cite{Zhang2016UnderstandingDL} demonstrated that DNNs are able to fit data with random labels and pure noise inputs. They raised the interesting question whether similar memorization tactics are used when DNNs are trained on data with true labels. \cite{arpit2017closer} showed that DNNs learn simple patterns first before memorizing, based on observations in extensive experiments. How DNNs prioritize the learning of simple patterns in early training has not been well-understood.

Paradigmatic examples of memorization algorithms are Lookup Tables and $k$-nearest neighbor approaches \cite{fix1951discriminatory}. \cite{chatterjee18a} built a network of Lookup Table Network that memorizes training examples. Although they memorize training examples, the built network also showed competitive generalization ability. Moreover, their built network shows little variation from run to run. \cite{cohen2018dnn} experimentally demonstrated the similarity between DNNs and $k$-NN.

It is evident that, given training examples and corresponding hyperparameters, the built $k$-NN model is unique. However, it is not clear whether the learned functions are also similar in the case of DNNs, when trained with different initializations, network sizes, and architectures. To explore this question, we propose an approach to measure the similarity and dissimilarity between what individual DNNs have learned and what they have memorized. By using our proposed approach, we find that DNNs have \textit{One way to Learn} and \textit{N ways to Memorize} training data. A deeper analysis shows that there is a qualitative difference between neural networks and $k$-NN.

Our contributions can be summarized as follows: 1) With extensive experiments, we show that how DNNs initially learn or prioritize simple patterns; 2) We propose an approach to measure the similarity between what different DNNs have learned and memorized given the same training data; 3) With the proposed approach, we analyze DNNs trained on data with true labels and random labels, respectively. We discuss our understanding of the dissection results.

\begin{figure}
     \centering
     \begin{subfigure}[b]{0.45\textwidth}
         \centering
         \includegraphics[scale=0.25]{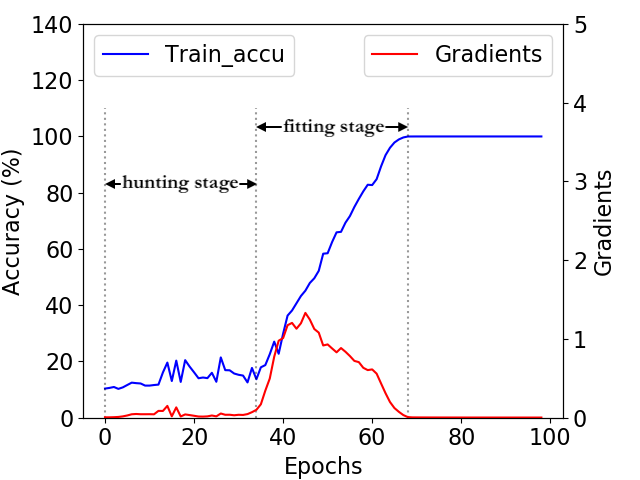}
         \caption{VGG11 trained on data with Random Labels}
         \label{fig:vgg11R}
     \end{subfigure}
     \begin{subfigure}[b]{0.45\textwidth}
         \centering
         \includegraphics[scale=0.25]{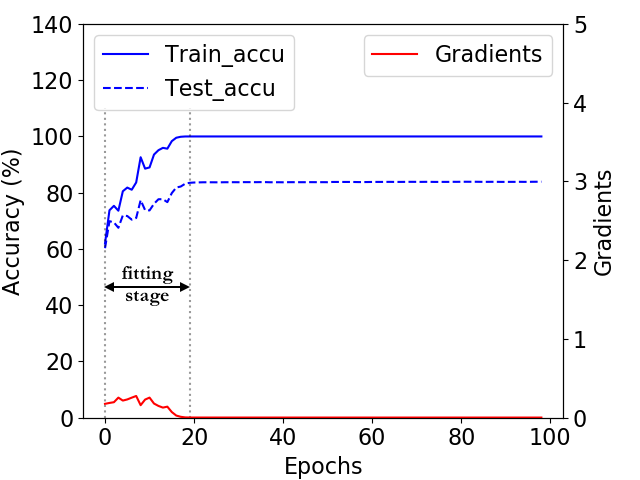}
         \caption{VGG11 trained on data with True Labels}
         \label{fig:vgg11T}
     \end{subfigure}
        \caption{The training and test accuracy, and the gradient magnitude $\overline{G}$ in the training processes.}
        \label{fig:vgg11}
\end{figure}

\vspace{-0.1cm}
\section{How DNNs Prioritize the Learning of Simple Patterns?}
\label{sec:prio}
\vspace{-0.1cm}
We first train VGG11 on CIFAR10 with true labels for 100 epochs using SGD with a constant learning rate of 0.01, moment 0.9 and a batch size of 128, using weight decay ($\lambda=$1e-4) as a regularizer. The model achieves 100\% accuracy on training data and 83,73\% accuracy on test data. Under the same setting, we train another VGG11 model with random labels without the regularizer, which achieves 100\% accuracy on training data and $\approx$10\% accuracy on test data.

Given the training loss $\mathcal{L}$ and a training data point $\pmb{x}=\{x_1, x_2, \cdots, x_M\}$, the gradient of the loss with respect to each input dimension is {\scriptsize $\frac{\partial \mathcal{L}}{\partial x_j}$}. The gradient magnitude of the sample is defined as the sum of absolute gradient values of an input {\scriptsize $G(\pmb{x}) = \sum^M_{j=1} |\frac{\partial \mathcal{L}}{\partial x_j}|$}. We record {\scriptsize $G(\pmb{x})$} during training and average them over all training examples {\scriptsize $\overline{G} = \frac{1}{N} \sum^N_{i=1} G(\pmb{x}^{(i)})$}.

In the computational graph of a DNN, inputs and parameters of the DNN are leaf nodes. Their received gradients depend on the gradients of shared intermediate nodes due to the chain rule. In a backpropagation, gradients become smaller and smaller when propagated back to the input layer. To a great extent, a big gradient magnitude $\overline{G}$ corresponds to large gradients of the parameters. 

Figure \ref{fig:vgg11} shows the accuracy and the gradient magnitude $\overline{G}$ of each epoch in the training process. In the model trained with random labels, the training process started with a hunting stage. Once it starts to fit, it converges quickly. This phenomenon is also observed in \cite{Zhang2016UnderstandingDL}. In the model trained with true labels, the model fits the data directly without a hunting stage. What we observed and summarised is not specific for the VGG11 model, but also for other models, e.g., a similar model with different network size (VGG13) and a model with a different architecture (Resnet32) (see Supplementary material).

We use a constant learning rate in our experiments. In the hunting stage, the model updates itself slowly since gradients are small. In the fitting stage, the big gradients make the quick update of the model possible to achieve high accuracy. An interesting question is how the gradients of parameters relate to the complexity of patterns in the training data.

In the training data with true labels, simple patterns (i.e., easily recognizable patterns shared by the images with of the same class) exist. Since visually similar images with similar intermediate feature representations have the same labels, the gradients of a parameter inside a batch are consistent, which leads to a quick update of the parameter. In contrast, in the training process on random labels, in the beginning, the corresponding gradients are partly canceled due to the inconsistent gradient information. After fine-tuning for several epochs, the neural network is able to map inputs of the same class (i.e., visually different images) to similar intermediate feature representations, which makes the gradients consistent again. The neural network starts the fitting stage.

\begin{figure}
\centering
\begin{minipage}{.46\textwidth}
  \centering
  \includegraphics[scale=0.45]{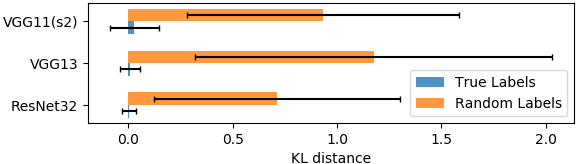}
	\caption{The dissimilarity between the function learned by VGG11(s1) and the functions learned by other models (VGG11(s2), VGG13, ResNet32): the thin blue bars correspond to the case where all models were trained with true labels; the thick orange bars correspond to models trained with random labels.}
	\label{fig:dissim}
\end{minipage}%
\hspace{0.3cm}
\begin{minipage}{.48\textwidth}
  \centering
  \includegraphics[scale=0.425]{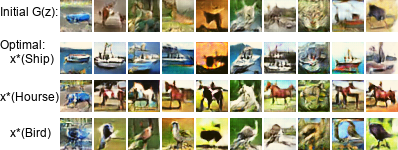}
	\caption{The images generated by the GAN: the first row shows the images generated by GAN with different random seeds; other rows show the ones after the optimization of Equation \ref{equ:disect_p} for different output classes of VGG11, such as \textit{Ship}, \textit{Hourse} and \textit{Bird}.}
	\label{fig:gen_img}
\end{minipage}
\end{figure}

\vspace{-0.2cm}
\section{Dissection Approach}
\vspace{-0.1cm}
One path to understand DNNs is to visualize what each neuron has learned or detected. Activation Maximization synthesizes an input that highly activates a specific neuron using the gradients of the neural activation with respect to the input \cite{erhan2009visualizing}. Different regularizations are applied \cite{yosinski2015understanding,mahendran2016visualizing,nguyen2016synthesizing} to make the synthesized inputs (i.e., images) look realistic. In this work, we aim to study the similarity between what different DNNs have learned and memorized given the same training data.

The two DNNs are trained on the same training data $\mathbb{X} = \{\pmb{x}^{(1)}, \pmb{x}^{(2)}, \cdots, \pmb{x}^{(n)}\}$ assumed to be sampled from a data distribution $\chi$. Their learned functions are $f^1$ and $f^2$, respectively. We can find the input pattern that activates $j$-th output class of $f^1$ using Activation Maximization \cite{erhan2009visualizing}, namely, $\pmb{x}^{j*} =\mathop{\argmax}_{\tiny \pmb{x}} f^1_j(\pmb{x})$. We could test whether $f^2$ also has a high response on the pattern at the $j$-th class $f^2_j(\pmb{x}^{j*})$. By dissecting $f^2$ with the patterns learned by $f^1$, we can compare the two functions.

However, the challenge is that $\pmb{x}^{j*}$ could be very unlikely considering the data distribution $\chi$. The $f^2$ then could show unexpected outputs on the out-of-distribution inputs, which makes the comparison not accurate. To overcome the challenge, we apply a Generative Adversary Network (GAN) \cite{goodfellow2014generative} to regularize the Activation Maximization process.

A GAN consists of two components: a Discriminator to distinguish the real data from generated data and a Generator to fool the Discriminator by generating data. In the case of convergence, the Generator is able to map latent variables to samples in data space. We train a GAN on the same training data $\mathbb{X}$ as above until convergence. 
The Generator therefore maps latent variables $\pmb{z} \sim \mathcal{N}(0,\, 1)$ to a sample $G(\pmb{z}) \sim \chi$. The pattern learned by the $j$-th output class of the DNN $f^1$ is
\begin{equation}
\footnotesize
\pmb{z}^{j*} = \mathop{\argmax}_{\pmb{z}} f^1_j(G(\pmb{z})) \qquad  \qquad  \pmb{x}^{j*} = G(\pmb{z}^{j*})
\label{equ:disect_p}
\end{equation}
where pre-trained $f^1$ and $G(\cdot)$ are fixed. The created $\pmb{x}^{j*}$ subject to $\chi$ can be used to dissect $f^2$.

Two strategies DNNs use are the lerning of simple patterns or the memorizing training samples. The pattern $\pmb{x}^{j*}$ in Equation \ref{equ:disect_p} is what the DNN $f^1$ has learned or memorized in the $j$-th class. By dissecting $f^2$ using the pattern, we can identify whether the two functions are similar. The similarity between $f^1$ and $f^2$ can be measured with the following metric.
\begin{equation}
\footnotesize
Dist(f^1, f^2) = \frac{1}{C} \sum_j^C D_{{\tiny KL}}(f^1(\pmb{x}^{j*})\| f^2(\pmb{x}^{j*}))
\label{equ:metric}
\end{equation}
The lower the dissimilarity value {\footnotesize $Dist(f^1, f^2)$} is, the more similar the two functions are. Another closely related work also uses a similar metric by replacing $\pmb{x}^{j*}$ with the images in test dataset \cite{hacohen2019all}. Different from their work, we focus on the patterns that DNNs has learned or memorized, not the random ones on the data distribution.

\vspace{-0.1cm}
\section{Dissecting DNNs via Experiments}
\vspace{-0.1cm}
The CIFAR10 training dataset with true labels is notated as $\mathbb{X}^T$. By replacing the true labels with random labels, we get another training dataset $\mathbb{X}^R$. We train four DNNs on $\mathbb{X}^T$ and $\mathbb{X}^R$, respectively. The first two models  VGG11(s1), VGG11(s2) are VGG11 trained with different random initializations (i.e., different seeds). The other two models have a different network size or architecture, namely, VGG13 and ResNet32. The setting of training processes is the same as in Section \ref{sec:prio}. They all achieve 100\% accuracy on the training datasets.

We train a deep convolutional GAN on the $\mathbb{X}^T$. The architectures and hyperparameters are the same as in \cite{Radford2015UnsupervisedRL}. The generator maps 100-dimensional gaussian noises $\pmb{z}$ to images $G(\pmb{z})$ with shape (32, 32, 3) using four transposed convolutional layers; the discriminator maps the images to a single output using four convolutional layers. They are trained on $\mathbb{X}^T$ for 30 epochs to converge.

\begin{figure}
\centering
\begin{minipage}{.48\textwidth}
  \centering
     \includegraphics[scale=0.28]{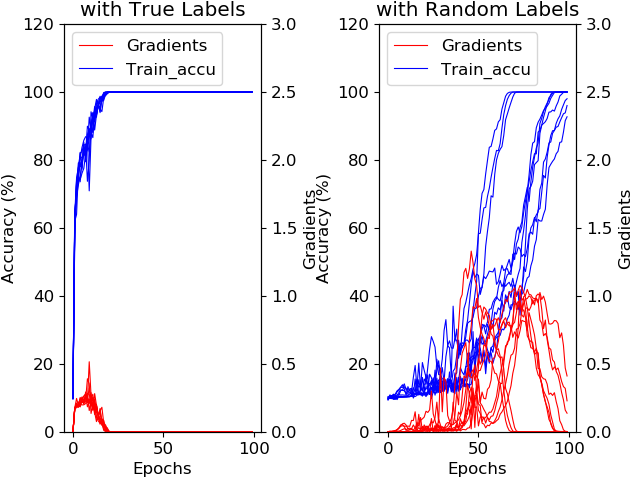}
	\caption{The model VGG11 is trained with 10 different seeds. Shown are training accuracy and gradient magnitude.}
	\label{fig:seeds}
\end{minipage}%
\hspace{0.2cm}
\begin{minipage}{.48\textwidth}
  \centering
  \includegraphics[scale=0.265]{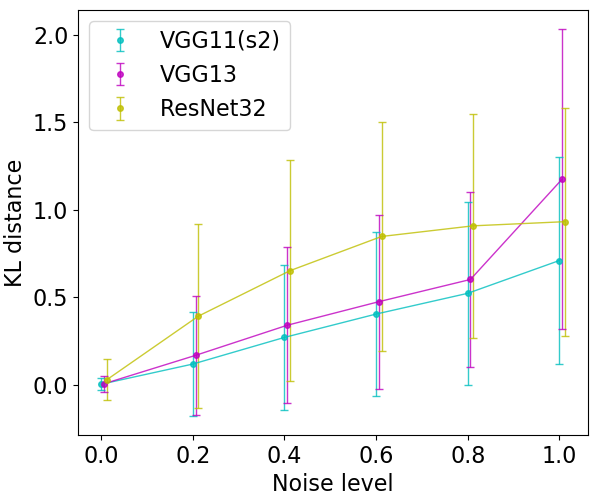}
  \caption{The similarity between models trained on data with different noise levels. With less noise, models have a greater similarity.}
  \label{fig:noise}
\end{minipage}
\end{figure}

Given the four models pre-trained on $\mathbb{X}^T$ or $\mathbb{X}^R$, we set VGG11(s1) model as $f^1$ and the rest of models as $f^2$, respectively. To compute $Dis(f^1, f^2)$ in Equation \ref{equ:metric}, for each class, we initialize $\pmb{z} \sim \mathcal{N}(0,\, 1)$ with 100 different seeds. The Equation \ref{equ:disect_p} is optimized by Gradient Ascent with a learning rate of 0.05 for 1000 iterations.

\textbf{One way to Learn:}
For well-trained models, the dissimilarity values between them are computed with Equation \ref{equ:metric} and shown in Figure \ref{fig:dissim}. Each bar corresponds to a averaged dissimilarity value between two models. The black line on each bar corresponds its variance.

The blue bars with very low values correspond to the dissimilarity between models trained on $\mathbb{X}^T$. What different models have learned are extremely similar. Given the training data $\mathbb{X}^T$, what neural networks have learned are not affected by different random initializations, network sizes, and neural architectures. In other words, DNNs have one way to learn.

We visualize the patterns what VGG11(s1) has learned. Given the $j$-th class, $\pmb{z}$ is initialized with different seeds. We show the images $G(\pmb{z})$ generated by GAN before the optimization of Equation \ref{equ:disect_p} and also the ones after the optimization $\pmb{x}^{j*}$. In Figure \ref{fig:gen_img},  each column corresponds to a seed, and the images in the 2nd-4th row correspond to the patterns learned for the respective classes.

\textbf{N ways to Memorize:}
As visible in Fig \ref{fig:dissim}, models trained on $\mathbb{X}^R$ memorize all the training samples. The yellow bars with high values mean that what different models have memorized from $\mathbb{X}^R$ are different, even between the models from run to run. DNNs have different ways to memorize samples.

In our experiments, as in previous works \cite{Zhang2016UnderstandingDL,arpit2017closer}, weight decay and data augmentations are turned off to reach zero error on the training data. We train VGG11 on $\mathbb{X}^R$ with the weight decay ($\lambda=$1e-4) and the standard data augmentation (4-pixels paddings, 32$\times$32 random croppings, and horizontal flips), respectively. The experiment results show that the weight decay and the data augmentation do not affect the similarity between what different models have memorized.

To further confirm our hypothesis about the qualitative difference between models trained on $\mathbb{X}^T$ and $\mathbb{X}^R$, we train VGG11 with 10 random seeds on the two datasets, respectively. We visualize the training accuracy and the gradient magnitude $\overline{G}$ of the 10 training processes in Figure \ref{fig:seeds}. For the models trained on $\mathbb{X}^T$, the optimization always follows similar paths, as shown in the left subplot. Contrarily, the right subplot shows that the training process on $\mathbb{X}^R$ follows obviously different optimization paths. We conjecture that the optimization on $\mathbb{X}^R$ encounters a large number of saddle points \cite{dauphin2014identifying,choromanska2015loss} and that the number of total saddle points encountered during learning depends on how the optimizer choose path directions on each saddle point, which might be quite sensitive to data.

\textbf{Learn and Memorize:}
When we train models on the training data with different noise levels (i.e., a certain percentage of training data are specified with random labels), it is not clear what DNNs learn and/or memorize. Our analysis in Section \ref{sec:prio} supports the hypotheses: instead of only learning on the simple patterns or being totally misled by the examples with random labels, DNNs start to learn simple patterns and then memorize the rest of samples. The learned functions become gradually far away from each other when the noise level increases (see Figure \ref{fig:noise}).

\vspace{-0.3cm}
\section{Conclusion}
\vspace{-0.2cm}
Our work builds a connection between the learning process and the complexity of patterns in training data by analyzing their gradients. We further propose an approach to compare what different DNNs learned and memorized. The experimental results reveal that DNNs have one way to learn and N ways to memorize.


\bibliography{neurips_2019}
\bibliographystyle{unsrt}

\end{document}